%% file: paper_pose_est.tex
\documentclass{article}
\usepackage{spconf,amsmath,graphicx}
\usepackage{times}
\usepackage{epsfig}
\usepackage{amssymb}
\usepackage{tabularx}

\usepackage[pagebackref=true,breaklinks=true,letterpaper=true,colorlinks,bookmarks=false]{hyperref}

\usepackage{setspace}

\title{Object 6D Pose Estimation with Non-local Attention}
\name{Jianhan Mei, Henghui Ding, Xudong Jiang}
\address{Nanyang Technological University, Singapore}
\include{definitions}

\begin{document}
%
\maketitle

\begin{abstract}
In this paper, we address the challenging task of estimating 6D object pose from a single RGB image. Motivated by the deep learning based object detection methods, we propose a concise and efficient network that integrate 6D object pose parameter estimation into the object detection framework. Furthermore, for more robust estimation to occlusion, a non-local self-attention module is introduced. The experimental results show that the proposed method reaches the state-of-the-art performance on the YCB-video and the Linemod datasets.
\end{abstract}

\section{Introduction}
\label{sec:introduction}

The estimation of object instance 6D pose has been a fundamental component in many application fields, \eg robotic manipulation and autonomous driving. It is challenging to estimate 6D object instance pose from 2D images since object information is lost during the projection from 3D to 2D.

Typically, the task of 6D object pose estimation is separated into two-stage: 1) instance object detection and 2) parameter estimation. Driven by the great success of deep learning~\cite{shuai2018toward, ding2018context, ding2020semantic, mei2019deepdeblur}, the object detection has been well-studied in recent years and lots of deep learning based methods, \eg \cite{FastRCNN, FasterRCNN, SSD, YOLO, YOLOV2}, have been proposed that achieved excellent performance in many scenarios. All of these methods crop the interested regions from the feature maps of the convolution layers and normalize them into a fixed size for the classification and the regression. The classifier and the regressor predict the object information by the local part of the feature maps. And the global information can be recovered by an inverse normalization. However, since 2D space normalization is not equivalent to 3D targets, the regressor suffers from loss of global information when directly integrating the 6D pose parameter into the framework of detection.

For the 6D parameter regression, the traditional methods focus on recovering the pose by matching key point features between 3D models and images ~\cite{MOPEDAlvaro, SIFTLowe, LocalAffineFred}. Such methods suffer the problem of key points extraction and description. The existing RGB-D data-based methods ~\cite{ModelTextureStefan, Learning6D3DObjCoorEric, LearningHierarchicalSparseLiefeng, RGBDPretrainedMax, DLLocalPatchWadim} improve the pose parameter regression significantly by using additional depth information. However, depth cameras have highly constrained configurations and are unavailable in some scenarios (\eg, outdoor scene). Hence in this paper, we address the problem of 6D parameter regression from RGB images, which is much easier to be obtained. Recently, PoseCNN ~\cite{PoseCNN} have shown that the object 6D pose information can be learned directly from the 2D image by utilizing the powerful learning capability of deep networks and the known camera intrinsic. However, the estimation networks still face problems of the parameter normalization, decoupling, and the prediction precision.

\begin{figure}[t]
    \centering
    \scalebox{1}{\includegraphics{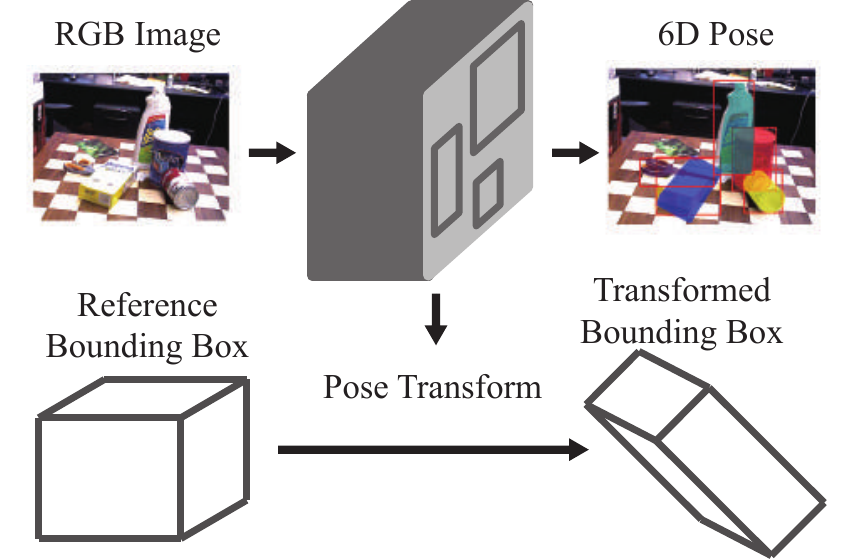}}
    \caption{The brief pipeline of the 6D object pose task.}
    \normalsize
    \label{fig:motivation}
\end{figure}

To build up a more efficient and effective network, we try to integrate 6D object pose parameter estimation into the object detection framework. Assuming the 6D object instance pose is defined by a reference 3D model, we use the 3D bounding box to map it to a unique 8 points 2D box. The 6D pose parameter is separated into the rotation and the translation which are normalized according to the Region of Interest (RoI) feature respectively. For better extraction of the object feature, a non-local based self-attention mechanism is introduced. By weighting the original feature using the non-local information, the final output features are more robust in case of the occlusion. For automatically finding the multi-task trade-off for rotation and translation parameters, the final loss is computered on the transformed 2D coordinates as shown in Fig.~\ref{fig:motivation}. For better illustration and fair comparison, we use Faster R-CNN (Region-based Convolutional Network)~\cite{FasterRCNN} as the backbone. Note that the proposed integration method is not limited to specific frameworks.

\section{Related Work}
\label{sec:related_work}

The acquisition of object instances information in an image is typically base on the detection approaches. Before the advent of the deep network based detection, Deformable Part Model (DPM) ~\cite{DPMPedro} and Selective Search ~\cite{SSJasper} are the most powerful detectors. In recent years, deep learning achieves great success~\cite{ding2019semantic, wang2019dermoscopic, liu2019feature, wang2019bi, ding2019boundary} and the deep learning based detection methods have made significant improvements. Start from the ~\cite{FastRCNN, FasterRCNN}, the two-stages frameworks are proposed and improved. Based on ~\cite{SPPKaiming}, the RoI Pooling is introduced in ~\cite{FasterRCNN} and used in the most recent two-stages detection frameworks. Before feeding the convolutional feature to the classifier, the feature is cropped and resized into fixed shape, which causes the information lost. Later, the single-shot detectors are proposed ~\cite{SSD, YOLO, YOLOV2}. The multi-scale bounding boxes attached with the convolutional feature maps are used instead of the proposals in the two-stages frameworks. The single-shot methods are typically faster than the two-stages framework while they still suffer the problem of the input resizing problem.

Understanding 3D scenes from 2D images has been studied for a long time. Especially for instance-level 3D parameter estimation, many research efforts have been tried ~\cite{3DRCNNAbhijit, Car3DBBox, FastSSDPoseEstPatrick, DataDriven3DYu, Seeing3DMathieu, AreCars3DMuhammad, Detailed3DZeeshan, 3DObjectDetectionandViewpointSanja}. The traditional pose estimation methods are roughly classified into the template-based methods and the feature-based methods. With the prevalence of the deep learning methods, deep neural network has been applied to the 6D object pose estimation task. Based on the object detection framework, either the key points or the pose parameters are regressed by the network. However, the deep learning based regression methods suffer the problem of the object occlusion. In~\cite{XiaolongNonLocal}, the non-local neural network is introduced. The non-local block is used as a self-attention, which enforce the network to use the global information. In this work, we try to utilize the self-attention property of the non-local neural network to have better object feature.

\section{Method}
\label{sec:method}
The task of recovering 6D pose parameters of all the object instances in a single RGB image consists of two parts, which are the object instance detection and the pose parameter regression. Based on one of the most popular two-stage object detection frameworks, Faster R-CNN, we integrate the 6D pose parameter regression into the deep network together with the object instance classification and localization. Further, the non-local neural block is introduced as a self-attention which makes the feature concentrate on its object parts and robust to the occlusion.

\subsection{Virtual RoI Camera Transform}
In ~\cite{3DRCNNAbhijit}, the allocentric and egocentric description problem of the global image and the proposal is discussed. ~\cite{3DRCNNAbhijit} uses the allocentric representation for learning parameter from the RoI features. The pose parameters of each object instance are re-defined by a canonical object center and a 2D amodal bounding box. By applying the perspective mapping, the global egocentric pose can be recovered. However, since the recovered egocentric pose is related to the predicted values, the canonical object center and the 2D amodal bounding box, the prediction errors from the translation and rotation parameters would interact with each other. Also in ~\cite{PoseCNN}, the importance of decoupling the regression of the translation and rotation parameters is claimed.

\begin{figure}[h]
    \centering
    \scalebox{1}{\includegraphics{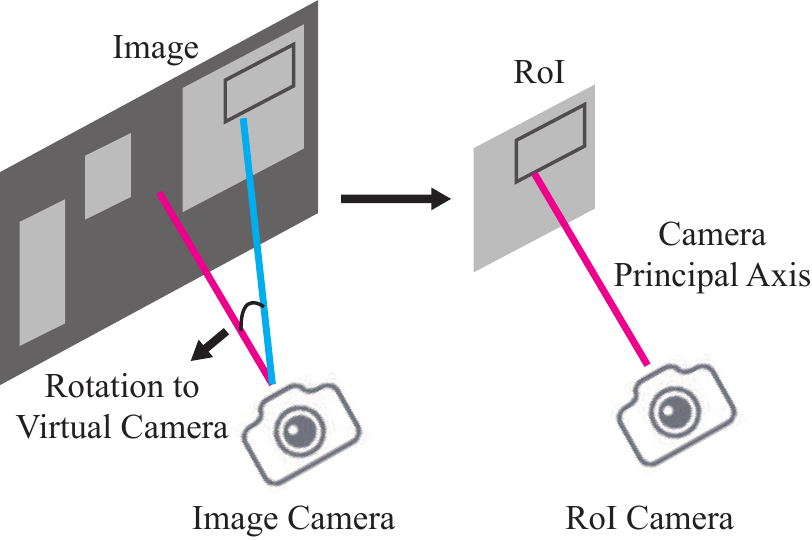}}
    \caption{Virtual RoI camera transform.}
    \normalsize
    \label{fig:virtual_cam}
\end{figure}

In this work, we normalize the 6D pose parameters according to the RoI proposals. In Faster R-CNN, the RoI features are cropped from the global feature map. Showing in Fig.~\ref{fig:virtual_cam}, as the description of the proposal, the cropped RoI feature can be regarded as a view changing transform of the original scene. Following ~\cite{MultipleViewGeometryAndrew, 3DRCNNAbhijit}, the virtual RoI camera and its intrinsic matrix is defined as:

\begin{equation}
{K_c} = \left[ {\begin{array}{*{20}{c}}
{{f_x}}&0&{{p_x}}\\
0&{{f_y}}&{{p_y}}\\
0&0&1
\end{array}} \right],{K_{RoI}} = \left[ {\begin{array}{*{20}{c}}
{{f_x}/{r_w}}&0&{0.5}\\
0&{{f_y}/{r_h}}&{0.5}\\
0&0&1
\end{array}} \right],
\label{equ:RoI_intrinsic_mat}
\end{equation}
where, $K_c$ is the camera intrinsic matrix, $K_{RoI\_norm}$ is the virtual RoI camera intrinsic matrix, $r_w$ and $r_h$ are width and height of the RoI.

Since that in a fixed network structure, the size of the RoI pooling is fixed. We normalize the camera changing mapping by the view of the virtual RoI camera $K_{RoI\_norm}$ and make each proposal mapped to a new coordinate space within $[0, 1]$. The virtual camera principal axis $[0, 0, 1]^T$ is mapped to the center point $(0.5, 0.5)$. According to ~\cite{MultipleViewGeometryAndrew}, the infinite homography matrix ${K_{RoI\_norm}}R_{RoI}^{ - 1}K_c^{ - 1}$ is used to define the 2D transformation between the virtual RoI view and the original image view, where $R_{RoI}$ is the rotation matrix from the image camera to the RoI camera. The 6D poses of the objects in each proposal are normalized by the virtual camera principal axis. Here, we discard the allocentric concept. The object poses both in the global image and in the RoI are considered as under the egocentric representation, but with a 2D transformation by the infinite homography matrix.

In Faster R-CNN, the object bounding boxes are normalized by the RoIs in Eq.~\ref{equ:RoI_coor_norm}:

\begin{equation}
\begin{array}{l}
{t_x} = (x - {x_a})/{r_w},{t_y} = (y - {y_a})/{r_h},\\
{t_w} = \log (w/{r_w}),{t_h} = \log (h/{r_h}),
\end{array}
\label{equ:RoI_coor_norm}
\end{equation}
where, $x, y, w, h$ are the left top coordinate and the width and the height of the object on the image, $x_a, y_a, r_w, r_h$ are the left top coordinate and the width and the height of the RoI on the image, $t_x, t_y, t_w, t_h$ are the left top coordinate and the width and the height of the object on the RoI.

Essentially, the object bounding boxes normalization in ~\cite{FastRCNN, FasterRCNN} can be regarded as the camera view changing without considering the 3D rotation so that the coordinates can be normalized simply by the width and height ratios of each RoI.

\subsubsection{Rotation Normalization}
As described in the previous section, the views of RoI proposals are transformed from the original image view by the infinite homography matrix. The rotation matrix $R_{RoI}$ defines the rotation from the image camera principal axis to the RoI proposal camera principal axis. The 6D object rotation parameter needs to be normalized by the RoI proposal camera principal axis. The center, $c_{RoI}=[x_{c\_RoI}, y_{c\_RoI}, 1]$, of the RoI proposal is used to calculate the RoI camera principal axis. The rotation matrix $R_{RoI}$ can be obtained by Rodrigues rotation formula ~\cite{Rodrigues}:

\begin{equation}
{R_{RoI}} = I + {c_k} \times {c_{RoI}} + \frac{{{{({c_k} \times {c_{RoI}})}^2}}}{{(1 + {c_k} \cdot {c_{RoI}})}},
\label{equ:rodrigues}
\end{equation}
where, $I$ is the 3 by 3 identity matrix, $\times$ and $\cdot$ denote the cross product and the inner product of vectors respectively.

During training the network, the rotation labels are represented as: ${R_{obj}} = {R_{RoI}R_{label}}$. And the rotation output is represented as the quaternion to constrain the rotation regression network output such that ${R_{pred}} \in SO(3)$. And the original object rotation can be recovered by ${R_{output}} = {R^{-1}_{RoI}R_{pred}}$.

\subsubsection{Translation and Depth Normalization}
The depth of an object cannot be directly reflected by only a single RGB image. Given a 3D translation vector $t=[x, y, d]$, we treat the 2D translation $[x, y]$ and $d$ separately.

Since all the cropped RoI features are resized to a fixed resolution, the depth $d$ cannot be estimated directly. The perspective method is used. The depth $d$ can be represented as:

\begin{equation}
{d_{obj}} = \log (\frac{{{m_I}}}{{{m_{RoI}}}})
\label{equ:depth_mapping}
\end{equation}
where, $m_I$ and $m_{RoI}$ are the area of the identity mapping 2D bounding box and the area of the RoI, $d_{label}$ is the label depth. The output depth can be recovered by ${d_{output}} = \frac{{{m_{RoI}}}}{{{m_I}\exp ({d_{pred}})}}$:

Similar to the rotation parameter, $R_{RoI}$ is used for final normalized 3D translation vector:

\begin{equation}
\begin{array}{l}
\begin{array}{l}
{[{x_{obj}},{y_{obj}},{d_{norm}}]^T} = {R_{RoI}}{[{x_{label}},{y_{label}},1]^T}\\
{t_{obj}} = {[{x_{obj}},{y_{obj}},{d_{obj}}]^T}
\end{array}
\end{array}
\label{equ:XT_trans_norm}
\end{equation}
where, the translation vector is first normalized by $R_{RoI}$. Then the depth is replaced by the objective depth $d_{obj}$. And $t_{obj}$ is the target object translation vector that will be further used for the parameter regression. And the original translation is recovered by using the inverse matrix $R_{RoI}^{-1}$.

\subsection{Non-local Attention}
The object pose estimation task often suffers from the problem of occlusion. We introduce the non-local self-attention~\cite{XiaolongNonLocal} to make the system concentrate on the non-occluded object part. The non-local image processing was first used in image filtering [~\cite{NonLocalMean}s]. In [~\cite{NonLocalMean}s], the algorithm calculates the filter response considering both local and distant pixels.

In ~\cite{XiaolongNonLocal}, the non-local mean operation is introduced in deep neural networks as:

\begin{equation}
{y_i} = \frac{1}{{C(x)}}\sum\limits_{\forall j} {f({x_i},{x_j})g({x_j})}
\label{equ:non_local_mean}
\end{equation}
where, $y$ is the output of the operation while $x$ is the input. $i$ denotes the output position and $j$ enumerates all position used for the non-local calculation. The function $f$ computes the scalar for its two inputs and $g$ maps $x$ to a new representation. The two functions can be of specific instantiations. $C(x)$ is a normalization factor.

Following ~\cite{XiaolongNonLocal}, we introduce a non-local block after the RoI pooling. The RoI features have fixed spatial size. As in Eq.~\ref{equ:non_local_mean}, each channel of a RoI feature map predicts the full-size attention mask. And the output of the block is a weighted feature for better spatial attention of the further tasks.

\begin{figure*}[t]
    \centering
    \scalebox{1}{\includegraphics{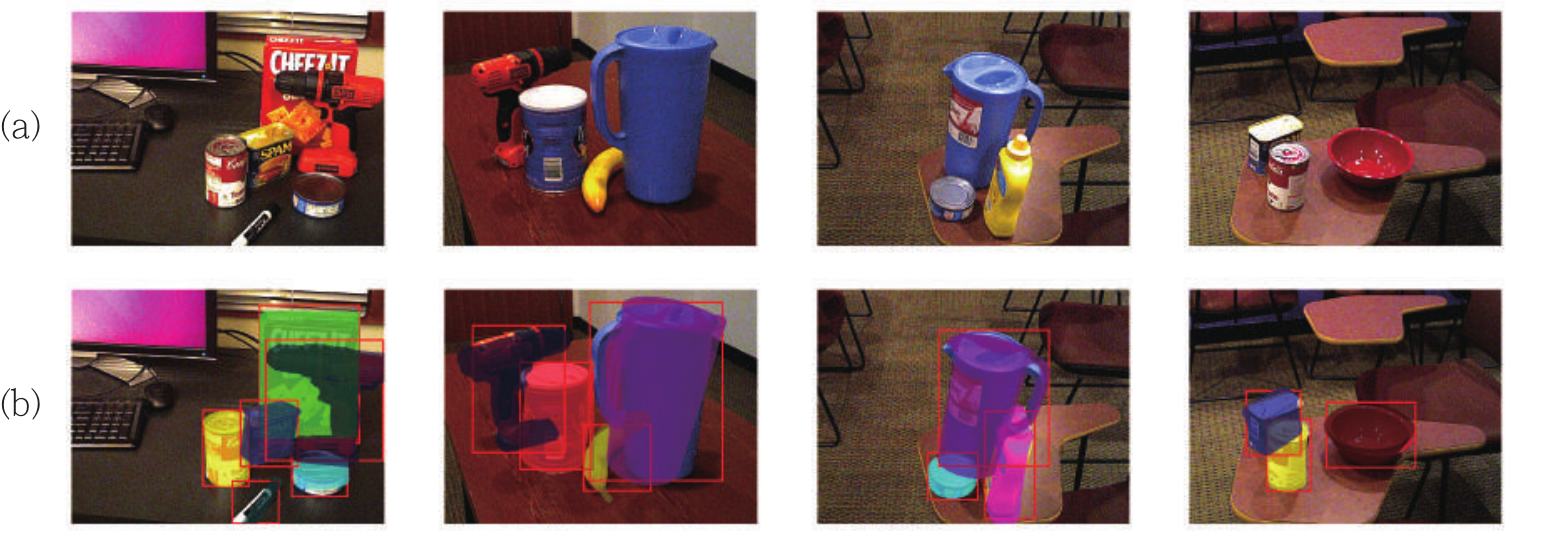}}
    \caption{(a) the original RGB images. (b) the images with transformed 3D models to 2D silhouette}
    \normalsize
    \label{fig:vis}
\end{figure*}

\subsection{Loss}
As the pose parameters are separated into translation and rotation, for better multi-task weighting, we do not directly regress the two parameters. The final parameter regression loss is computed on the transformed coordinates. More specifically the smooth L1 function~\cite{FasterRCNN} is used for all the parameter regression.

\begin{equation}
{L_{coordinate}} = {\left\| {{T_{pred}}(p) - {T_{label}}(p)} \right\|_{smooth\_L1}}
\label{equ:smooth_L1}
\end{equation}
where, $T(\cdot)$ denotes the transform $T$ to its point coordinates $p$. In the 6D pose case, $T$ can be represented as transform matrix form:

\begin{equation}
T = \left[ {\begin{array}{*{20}{c}}
R&t\\
0&1
\end{array}} \right]
\label{equ:T_matrix}
\end{equation}
where, $R$ is the rotation matrix, $t$ denotes the translation vector. The last row is the homogeneous extension.

\section{Experiment}
\label{sec:experiment}

\subsection{Dataset and Setup}
Our experiments are mainly conducted on the YCB-Video~\cite{PoseCNN} dataset. For YCB-Video dataset, following dataset split in ~\cite{PoseCNN}, the 80 videos with 80,000 synthetic images are for training and 2,949 key frames are for testing. For evaluation metrics, the average distance (ADD) and ADD symmetry (ADD-S) are tested. For testing the robust of the network, the Linemod dataset~\cite{ModelTextureStefan} is evaluated following the settings of~\cite{EricUncertainty}.

The model is implemented using the TensorFlow library. The VGG16 network is initialized by the pre-trained model on ImageNet. And the Adam optimizer is used for gradient calculation and weights updating.

\subsection{Results on the YCB-Video Dataset}
Following~\cite{PoseCNN}, the 3D coordinate regression network is chosen as the baseline network. And the single RGB version of PoseCNN~\cite{PoseCNN} is compared with the propose method.

\begin{table}[h]
    \small
	\centering
	\caption{Performance evaluation on the YCB-Video dataset}
	\begin{tabular}{p{0.98cm}<{\centering}|p{1.4cm}<{\centering}|p{1.2cm}<{\centering}|p{1.1cm}<{\centering}|p{1.6cm}<{\centering}}
	\hline
	Method &\tabincell{c}{3D \\ Coordinate} &\tabincell{c}{PoseCNN} &\tabincell{c}{Proposed} &\tabincell{c}{Proposed with \\ self-attention} \\
	\hline
	\hline
	ADD & 15.1 &53.7 &50.3 &\textbf{53.9} \\
	\hline
	ADD-S &29.8 &75.9 &75.1 &\textbf{77.0} \\
	\hline
	\hline
	\end{tabular}
	\label{Tab:PerformaceYCB}
\end{table}

As shown in Table~\ref{Tab:PerformaceYCB}, the proposed method without the non-local self-attention module reaches the comparable performance of PoseCNN~\cite{PoseCNN} where we do not use the symmetry distance loss and the Hough Voting based translation regression that needs an extra segmentation branch. Furthermore, with the self-attention module, our network outperforms the previous methods. Some of the visualized results are shown in Fig.~\ref{fig:vis}.

\subsection{Results on the Linemod Dataset}
The Linemod dataset~\cite{ModelTextureStefan} contains 15 objects. For each object, it has 1200 images. The objective item is annotated in each object image set. We use the training setting as~\cite{MahdiBB8}. The results are shown in Table~\ref{Tab:PerformaceLinemod}:

\begin{table}[h]
	\small
	\centering
	\caption{Performance evaluation on the Linemod Dataset}
	\begin{tabular}{p{1.2cm}<{\centering}|p{2cm}<{\centering}|p{2cm}<{\centering}}
	\hline
	Method &\tabincell{c}{PoseCNN} &\tabincell{c}{Proposed with \\ self-attention} \\
	\hline
	\hline
	Pose &62.7 &\textbf{64.3} \\
	\hline
	\hline
	\end{tabular}
	\label{Tab:PerformaceLinemod}
\end{table}

Consistent with the YCB-video result, our network outperforms PoseCNN on the evaluation metric of~\cite{EricUncertainty}, which demonstrates the robustness of the proposed network.

\section{Conclusion}
\label{sec:Conclusion}

In this work, we discuss the integration of 6D object pose estimation into the prevalent deep learning based object detection framework. The 6D pose parameter is separated into rotation and translation that are normalized separately. Moreover, a non-local self-attention mechanism is introduced to obtain better performance robust to occlusion. Experimental results demonstrate the feasibility of the proposed network, which reaches the state-of-the-art performance on the two datasets.


\begin{spacing}{0.95}
{
\footnotesize 
\bibliographystyle{IEEEbib}
\bibliography{ref}
}
\end{spacing}

\end{document}

%% file: definitions.tex
 \usepackage{xspace}

\newcommand{\tabincell}[2]{\begin{tabular}{@{}#1@{}}#2\end{tabular}}

\makeatletter
\DeclareRobustCommand\onedot{\futurelet\@let@token\@onedot}
\def\@onedot{\ifx\@let@token.\else.\null\fi\xspace}

\def\eg{\emph{e.g}\onedot}

\makeatother